\documentclass[runningheads]{llncs}

\usepackage{todonotes}
\usepackage{lipsum}
\usepackage{dsfont}
\usepackage{hyperref}
\usepackage{graphicx}
\usepackage{multirow}
\usepackage{multicol}
\usepackage{wrapfig}
\usepackage{subcaption}

\AtBeginDocument{%
  \let\mathbb\relax
  \DeclareMathAlphabet\PazoBB{U}{fplmbb}{m}{n}
  \newcommand{\mathbb}{\PazoBB}
}

\begin{document}
\title{Learning Visual Models using a \\ Knowledge Graph as a Trainer}

%
%

\author{Sebastian Monka\inst{1,2} \and
Lavdim Halilaj\inst{1} \and
Stefan Schmid\inst{1} \and
Achim Rettinger\inst{2}}
\authorrunning{S. Monka et al.}
\titlerunning{Learning Visual Models using a \\ Knowledge Graph as a Trainer}
%
\institute{Bosch Research, Renningen, Germany \\
\email{\{sebastian.monka,lavdim.halilaj,stefan.schmid\}@de.bosch.com} \and
Trier University, Trier, Germany \\
\email{\{rettinger\}@uni-trier.de}}

\maketitle              
%
\begin{abstract}
Traditional computer vision approaches, based on neural networks (NN), are typically trained on a large amount of image data.
By minimizing the cross-entropy loss between a prediction and a given class label, the NN and its visual embedding space are learned to fulfill a given task.
However, due to the sole dependence on the image data distribution of the training domain, these models tend to fail when applied to a target domain that differs from their source domain.
To learn a more robust NN to domain shifts, we propose the \emph{knowledge graph neural network} (KG-NN), a neuro-symbolic approach that supervises the training using image-data-invariant auxiliary knowledge.
The auxiliary knowledge is first encoded in a knowledge graph with respective concepts and their relationships, which is then transformed into a dense vector representation via an embedding method.
Using a contrastive loss function, KG-NN learns to adapt its visual embedding space and thus its weights according to the image-data invariant knowledge graph embedding space.
We evaluate KG-NN on visual transfer learning tasks for classification using the mini-ImageNet dataset and its derivatives, as well as road sign recognition datasets from Germany and China.
The results show that a visual model trained with a knowledge graph as a trainer outperforms a model trained with cross-entropy in all experiments, in particular when the domain gap increases.
Besides better performance and stronger robustness to domain shifts, these KG-NN adapts to multiple datasets and classes without suffering heavily from catastrophic forgetting.

\end{abstract}
\section{Introduction}
Deep neural networks (NNs) are widely used in computer vision (CV). 
Their main strength lies in their ability to find complex underlying features in images. 
A common method for training an NN is to minimize the cross-entropy loss, which is equivalent to maximizing the negative log-likelihood between the empirical distribution of the training set and the probability distribution defined by the model. 
This relies on the independent and identically distributed (i.i.d.) assumptions as underlying rules of basic machine learning, which state that the examples in each dataset are independent of each other, that train and test set are identically distributed and drawn from the same probability distribution~\cite{DBLP:books/daglib/0040158}.
However, if the train and test domains follow different image distributions the i.i.d. assumptions are violated, and deep learning leads to unpredictable and poor results~\cite{tan2018survey}.
This has been demonstrated by using adversarially constructed examples~\cite{Goodfellow2015ExplainingAH} or variations in the test images such as noise, blur, and JPEG compression~\cite{Hendrycks2019BenchmarkingNN}.
Authors in~\cite{DAmour2020UnderspecificationPC} even claim that any standard NN suffers from such an unpredictable distribution shift when it is deployed in the real world.

\begin{figure*}[!tb]
    \centering
    \begin{subfigure}[b]{0.27\textwidth}
        \centering
        \includegraphics[width=\textwidth]{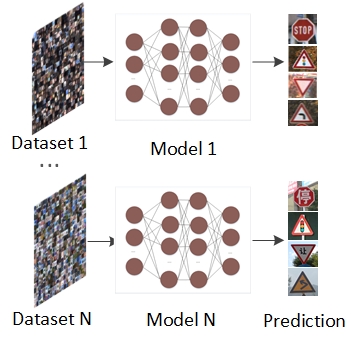}
        \caption{MDMM}
        \label{fig:current1}
    \end{subfigure}
    \hfill
    \begin{subfigure}[b]{0.34\textwidth}
        \centering
        \includegraphics[width=\textwidth]{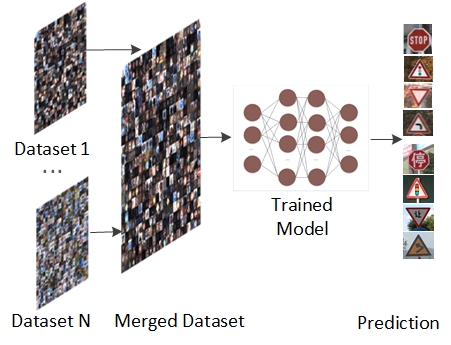}
       	\caption{SDSM}
	\label{fig:current2}
    \end{subfigure}
    \hfill
    \begin{subfigure}[b]{0.33\textwidth}
        \centering
        \includegraphics[width=\textwidth]{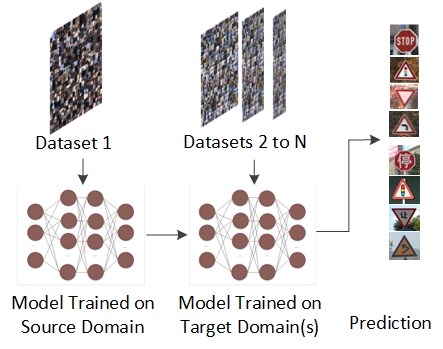}
        	\caption{MDSM}
	        \label{fig:current3}
    \end{subfigure}
    \caption{Categorization of domain adaptation approaches: a) Multiple Domains Multiple Models (MDMM); b) Single Domain Single Model (SDSM); and c) Multiple Domains Single Model (MDSM).}
    \label{fig:currentState}
\end{figure*}

Transfer learning approaches that deal with such distribution shifts can be grouped into three main categories as depicted in Figure~\ref{fig:currentState}: a) \emph{Multiple Domains Multiple Models (MDMM)}; b) \emph{Single Domain Single Model (SDSM)}; and c) \emph{Multiple Domains Single Model (MDSM)}.
MDMM approaches treat all datasets as independent and train a respective model for each of them.
Therefore, these approaches are very costly to train, and learned knowledge cannot be transferred between datasets.
SDSM approaches train a single model on a large dataset merged from many smaller ones. 
However, it is difficult to create a balanced dataset required by the NN to learn a general representation suitable for all domains.
MDSM approaches train a single model on various datasets at different stages, and can therefore transfer learned knowledge to new domains.
However, if trained with the standard cross entropy these models suffer from an unpredictable and error-prone knowledge transfer and \emph{catastrophic forgetting}, where learned knowledge from previous datasets tends to be forgotten after training on the current dataset.

To reduce the high dependency on the training domain, pre-training methods that generate rich embedding spaces seem to be a promising research direction for CV and natural language processing (NLP).
Exploring these embedding spaces, it is found that NNs encode visually similar classes close to each other when sufficient training data is available.
Recently, the idea of training an NN with an image-independent embedding space in form of language embeddings has also been proven to be beneficial for CV tasks~\cite{DBLP:conf/eccv/JoulinMJV16,radford2learning,DBLP:journals/corr/abs-2010-00747}.

In this paper, we introduce the \emph{knowledge graph neural network} (KG-NN), a novel approach to learn a visual model using a knowledge graph (KG) and its knowledge graph embedding $\vec{h}_{KG}$ as a trainer.
More concretely, a domain-invariant embedding space using a KG and an appropriate KG embedding algorithm is constructed.
We then train KG-NN with a contrastive loss function to adapt its visual embedding to $\vec{h}_{KG}$ given by the KG.
KG-NN, therefore, learns the relevant features of the images by connecting semantically similar classes and distinguishing them from different ones.
The benefit is two-fold.
First, KG-NN will be more robust to distribution shifts since the embedding space is independent of the dataset distribution, and second, it is enriched with additional semantic data in a controlled manner.

To investigate the generalization and adaption of KG-NN in real-world scenarios, the task of visual transfer learning provides a suitable testing environment.
Transfer learning tasks consist of a source and a target dataset, differing in terms of their underlying distribution, e.g sensors, environments, countries.
A domain generalization task has only access to labeled source data, whereas the domain adaptation task contains a small amount of additional labeled target data.
For domain generalization - \emph{Scenario 1}, we performed two experiments: 1) object classification, where the NN is trained on the mini-ImageNet~\cite{DBLP:conf/nips/VinyalsBLKW16} dataset and evaluated on derivatives; 2) road sign recognition, where the NN is trained on the German Traffic Signs Dataset (GTSRB)~\cite{Stallkamp2012ManVC} and evaluate on the Chinese Traffic Signs Dataset (CTSD)~\cite{Yang2016TowardsRT}.
For domain adaptation - \emph{Scenario 2}, we train the NN on GTSRB and additional labeled target data from CTSD.
In both scenarios, the respective KGs are developed in Resource Description Framework (RDF) representation.
RDF provides the necessary means for an easy and flexible extension of the defined schemas and allows for enriching and interlinking entities in the KGs with complementary information from other sources.

The generality of our approach becomes apparent in the fact that it can be assigned to any of the three categories illustrated in Figure~\ref{fig:currentState} since we provide an alternative and enriched training method for NNs.
While in this paper, we only compare with approaches from the third category, our results indicate that KG-NN is significantly more accurate compared to a conventional approach based on the cross-entropy loss in any domain-changing scenario.
Our main contributions of this paper are summarized as follows:
\begin{itemize}
    \item We introduce KG-NN, a neuro-symbolic approach that uses prior domain-invariant knowledge captured by a KG to train an NN.
    \item We adapt a contrastive loss function to combine knowledge graph embeddings with the visual embeddings learned by the NN.
    \item We evaluate the KG-NN approach in domain generalization and domain adaptation tasks on two different scenarios with respective image datasets.
\end{itemize}

The paper starts with the definition of preliminaries in Section~\ref{sec:preliminaries}.
Section~\ref{sec:knowledge graph as a trainer} presents a detailed description of KG-NN where a KG is used as a trainer.
Section~\ref{sec:evaluation} provides an evaluation on two datasets in a domain generalization and domain adaptation task.
Related work is outlined in Section~\ref{sec:related-work}.
We conclude the paper and provide an outlook on future directions in Section~\ref{sec:conclusion}.

\section{Preliminaries}
\label{sec:preliminaries}
\paragraph{Knowledge Graph.} 
We adopt the definition given by Hogan et al.~\cite{DBLP:journals/corr/abs-2003-02320} where a knowledge graph is \emph{a graph of data aiming to accumulate and convey real-world knowledge, where entities are represented by nodes and relationships between entities are represented by edges}.
In its most basic form, a KG is a set of triples $G = {H, R, T}$, where $H$ is a set of entities, $T \subseteq E \times L $, is a set of entities or literal values and $R$, a set of relationships which connects $H$ and $T$.

\paragraph{Knowledge Graph Embedding.} A knowledge graph embedding $\vec{h}_{KG}$ is a representation of entities and edges of a KG in a high-dimensional vector space while preserving its latent structure~\cite{DBLP:journals/corr/abs-2003-02320}.
Related to language embeddings, we count $\vec{h}_{KG}$ as a form of a semantic embedding $\vec{h}_{s}$.
The $\vec{h}_{KG}$ is learned by a knowledge graph embedding method $KGE(\cdot)$ using entities and relations encoded in the $KG$. 
Individual vectors, corresponding to the entities from the $KG$ represented in $\vec{h}_{KG}$ are denoted as $\vec{h}_{KG,a}$ with dimensionality $d_P$.

\paragraph{Visual Embedding.}
An \emph{encoder network} $E(\cdot)$ is part of the NN and maps images $\vec{x}$ to a visual embedding $\vec{h}_{v} = E(\vec{x}) \in \mathbb{R}^{d_E}$, where the activations of the final pooling layer and thus the representation layer have a dimensionality $d_E$, where $d_E$ depends on the encoder network that is used. If the encoder network is learned using a semantic embedding, we define it as $\vec{h}_{v(s)}$.
If the semantic embedding is given by a KG we further denote the visual-semantic embedding as $\vec{h}_{v(KG)}$.

\paragraph{Visual Projection.}
A \emph{projection network} $P(\cdot)$ maps the normalized embedding vectors $\vec{h}_{v}$ into a visual projection $\vec{z} = P(\vec{h}_{v}) \in \mathbb{R}^{d_P}$ in which it is compared with the class-label representation of the $\vec{h}_{KG}$. 
For the projection network $P(\cdot)$, we use a multi-layer perceptron \cite{DBLP:books/sp/HastieFT01} with a single hidden layer, an input dimensionality $d_E$, and output vector of size $d_P$ to match the dimensionality of $\vec{h}_{KG}$.

\paragraph{Transfer Learning.}
A formal definition of transfer learning is presented in \cite{DBLP:conf/emnlp/RuderP17} as follows: \emph{Given a source domain $D_S$ with input data $X_S$, a corresponding source task $T_S$ with labels $Y_S$, as well as a target domain $D_T$ with input data $X_T$ and a target task $T_T$ with labels $Y_T$, the objective of transfer learning is to learn the target conditional probability distribution $P_T (Y_T | X_T )$ with the information gained from $D_S$ and $T_S$ where $D_S \neq D_T$ or $T_S \neq T_T$}.
Transfer learning with no target data at training is referred to as domain generalization, whereas supervised domain adaptation has access to a small amount of labeled target data.

\section{Knowledge Graph as a Trainer}
\label{sec:knowledge graph as a trainer}
In this section, we define the basic terminology of the KG-NN approach as well as the underlying pipeline for the realization of a transfer learning task.

\begin{figure*}[t]
\centering
\begin{subfigure}[b]{0.49\textwidth}
\centering
\includegraphics[width=\textwidth]{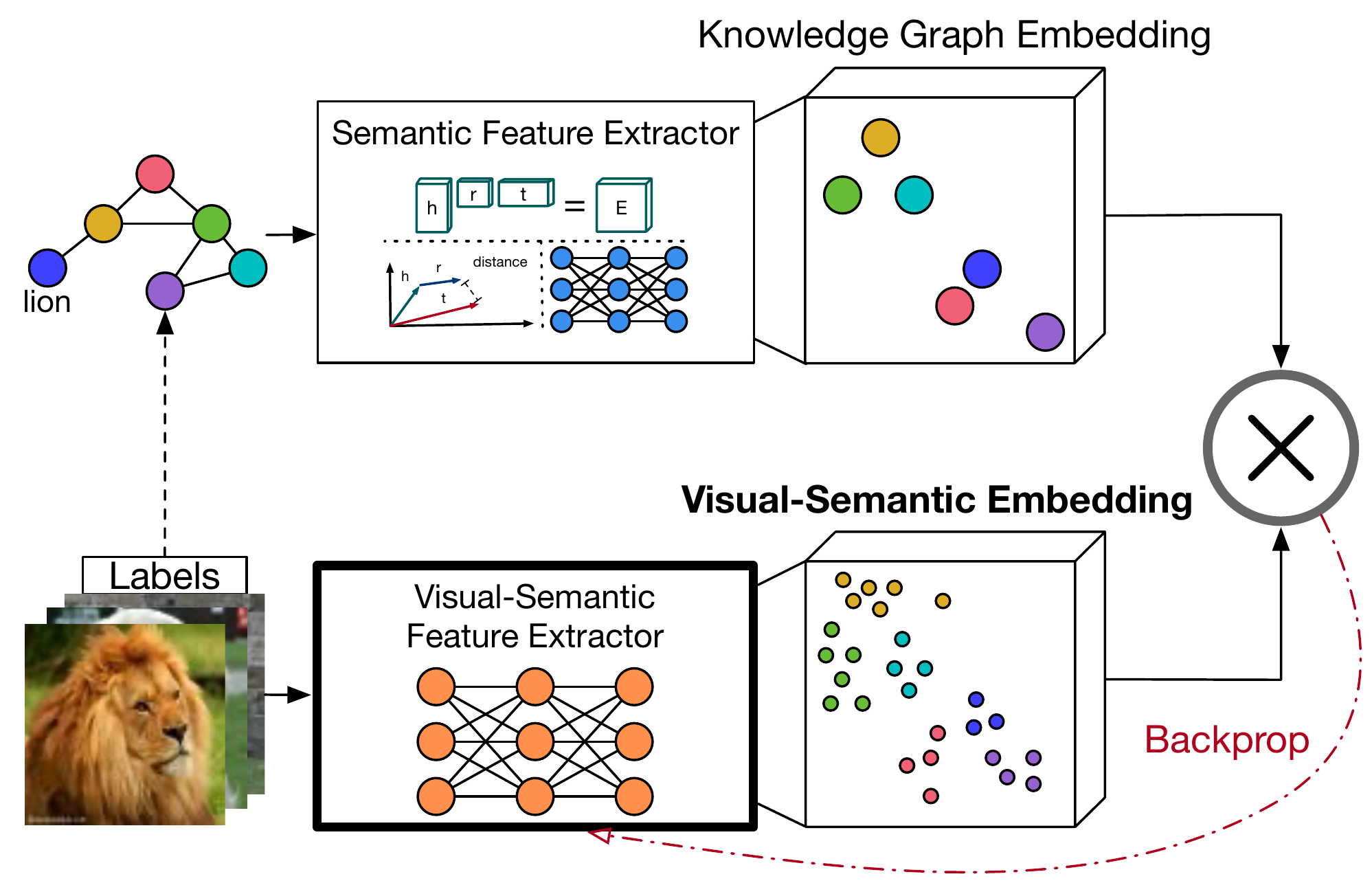}
\subcaption{Training abstraction of $\vec{h}_{v(KG)}$.}
\label{fig:Knowledge Graph as a Trainer overview}
\end{subfigure}
\begin{subfigure}[b]{0.49\textwidth}
\centering
\includegraphics[width=\textwidth]{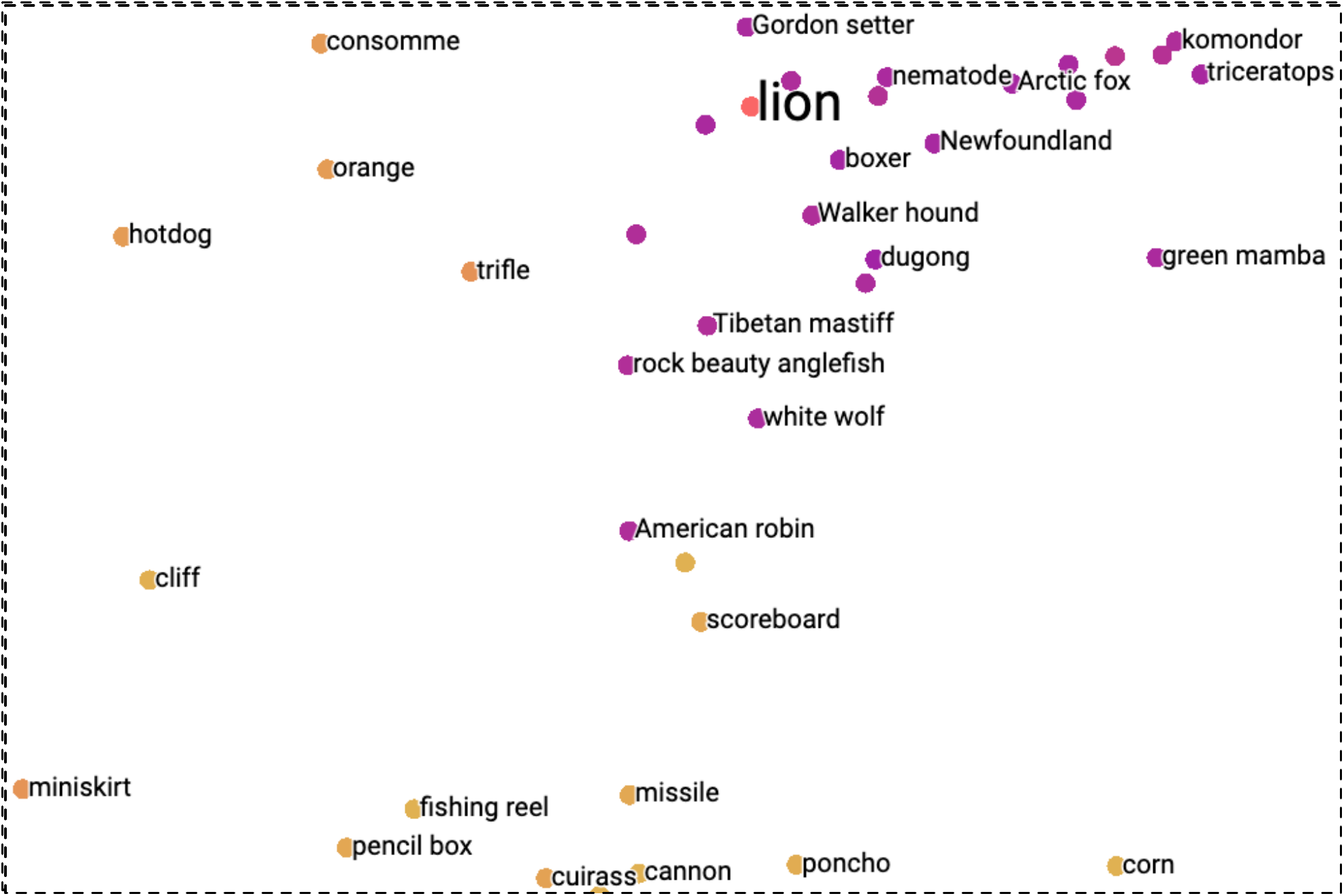}
\subcaption{Knowledge graph embedding $\vec{h}_{KG}$.}
\label{fig:Knowledge Graph as a Trainer semantic}
\end{subfigure}
\caption{KG-NN Approach: a) the main building blocks for learning a visual-semantic embedding space $\vec{h}_{v(KG)}$ using a knowledge graph as a trainer; b) the 2D projection of the semantic-embedding  $\vec{h}_{KG}$ represented in a knowledge graph.}
\label{fig:Knowledge Graph as a Trainer}
\end{figure*}

The main objective of KG-NN is incorporating prior knowledge into the deep learning pipeline using a knowledge graph as a trainer.
As depicted in Figure~\ref{fig:Knowledge Graph as a Trainer overview}, the class labels of a given dataset are infused to the NN in form of a high-dimensional vector of the knowledge graph embedding space $\vec{h}_{KG}$, instead of using the standard one-hot encoded vectors.
This $\vec{h}_{KG}$ shown in Figure~\ref{fig:Knowledge Graph as a Trainer semantic} is generated from a KG using a knowledge graph embedding method $KGE(\cdot)$.
It incorporates domain-invariant relations to other classes inside or outside the dataset and therefore enriches the NN with auxiliary knowledge in an indirect manner.
To guide the adaption of the NN to the $\vec{h}_{KG}$ space, we use the \emph{contrastive knowledge graph embedding loss}.
It compares the respective outputs from the visual feature extractor with the class label vectors of the $\vec{h}_{KG}$ forming a visual-sematic embedding space $\vec{h}_{v(KG)}$.
As a result, the learned NN projects respective images close to their representations given by the $\vec{h}_{KG}$.

\paragraph{Contrastive Knowledge Graph Embedding Loss.}
\label{subsec:ContrastiveKnowledgeGraphEmbeddingLoss}
We derive the contrastive knowledge graph embedding loss from the supervised contrastive loss~\cite{DBLP:conf/nips/KhoslaTWSTIMLK20,DBLP:conf/icml/ChenK0H20} which extend the multi-class N-pair loss~\cite{DBLP:conf/nips/Sohn16} or InfoNCE loss~\cite{DBLP:journals/corr/abs-1807-03748} with class label information.
Instead of contrasting images in the batch against an anchor image, we adapt the loss to contrast images of the batch against the class label representation of the $\vec{h}_{KG}$.
A batch consists of 2N training samples, two augmented versions for each of the N training images.
Within a batch, an anchor i $\in \{1...2N\}$ is selected that corresponds to a specific class label $\vec{y}_i$ and therefore assigns a specific embedding vector of the $\vec{h}_{KG}$, $h_{KG,i}$.
Positive samples are all samples that correspond to the same class label as the anchor i.
The numerator in the loss function computes a similarity score between the anchor vector of the $\vec{h}_{KG}$, $\vec{h}_{KG,i}$, and the visual projection vector of a positive sample in the batch, $\vec{z}_{j}$.
The denominator computes the similarity score between the anchor vector of the $\vec{h}_{KG}$ and the visual projection vector of all other samples $\vec{z}_{k}$ in the batch.
We choose the cosine similarity as the distance measure in the high-dimensional space. 
For each anchor $i$, there can be many positive samples, which contribute to the final loss, where $N_{\vec{y}_i}$ is their total number.
The KG-based contrastive loss function is then given by:
\begin{equation} 
    \mathcal{L}_{KG} = \sum^{2N}_{i=1} \mathcal{L}_{KG,i}
\label{eq:1}
\end{equation}
with
\begin{equation}
    \mathcal{L}_{KG,i} = \frac{-1}{2N_{\vec{y}_i} - 1} \sum_{j = 1}^{2N} \mathds{1}_{i \neq j} \cdot \mathds{1}_{\vec{y_i=\vec{y}_j}} \cdot \log \frac{\exp{(\vec{h}_{KG,i} \cdot \vec{z}_j / \tau)}}{\sum^{2N}_{k=1} \mathds{1}_{i \neq k} \exp{(\vec{h}_{KG,i} \cdot \vec{z}_k / \tau)}}
\label{eq:2}
\end{equation}
\noindent where $\vec{z}_l = P(E(\vec{x}))$, $\mathds{1}_{k \neq i} \in \{0, 1\}$ is an indicator function that returns 1 iff $k \neq i$ evaluates as true, and $\tau > 0$ is a predefined scalar temperature parameter.
During optimization of the loss function $ \mathcal{L}_{KG} $, the NN learns its weights by mapping its projection $\vec{z}_l$ to the $\vec{h}_{KG}$ space.

\begin{figure}[tb]
    \centering
     \small
    \includegraphics[width=\textwidth]{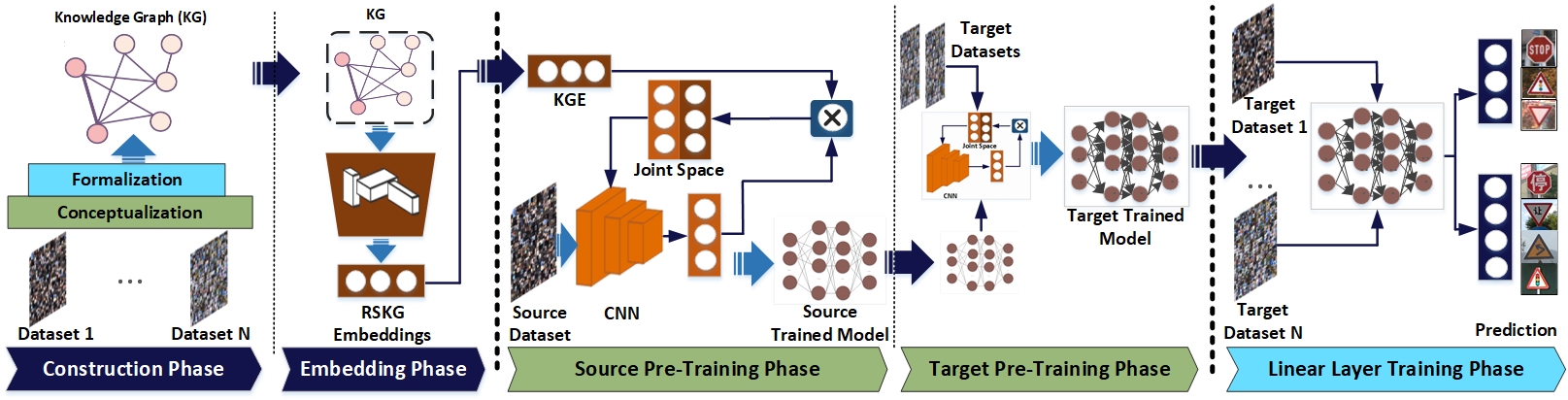}
    \caption{The designed pipeline consisting of five phases where a knowledge graph acts as a trainer supporting adaption and generalization: 
    \emph{Knowledge Graph Construction};
    \emph{Knowledge Graph Embedding}; 
    \emph{Source Domain Pre-Training};
    \emph{Target Domain Pre-Training};
    and \emph{Linear Layer Training}.}
    \label{fig:architecture}
\end{figure}

\subsection{Adaptation to a Labeled Target Domain}
\label{subsec:Adaptation to a Labeled Target Domain}

Training robust NNs is crucial in real-world scenarios as deployment domains typically differ from the training ones.
The knowledge graph as a trainer can influence how an NN should behave in different environments by providing a stable embedding space.
However, if the domain gap is quite large, it is beneficial to fine-tune the NN on labeled data of the target domain.
We design a training pipeline to support a transfer learning scenario where a small amount of labeled target data exists.
An overview of this pipeline comprised of five consecutive phases is shown in Figure~\ref{fig:architecture}.

\paragraph{Knowledge Graph Construction.}
Knowledge graphs can represent prior knowledge encoded with rich semantics in a graph structure.
Based on the selected scenario, underlying knowledge of one or multiple domains is conceptualized and formalized into a KG.
Since KGs are manually curated by human experts, it is possible to define an underlying schema comprising multiple classes from different domains.
This joint representation of domains enables inferring relations between classes, which can then be transferred into high-dimensional vector space.

\paragraph{Knowledge Graph Embedding.}
The KG is transformed into a knowledge graph embedding space $\vec{h}_{KG}$ via a knowledge graph embedding method $KGE(\cdot)$.
There are various approaches to generate these dense vectors that encode all entities and relations within the KG~\cite{Nickel2016HolographicEO,DBLP:conf/aaai/DettmersMS018,DBLP:conf/naacl/NguyenNNP18}.
Note that KG-NN can operate on any $\vec{h}_{KG}$ generated by any $KGE(\cdot)$, as an $\vec{h}_{KG}$ only reflects similarities between entities by distances and positions in the vector space.
Thus, if entities share many properties in the KG, they are closely located in space.
\paragraph{Source Domain Pre-Training.} 
We train KG-NN from scratch using the KG as a trainer and do not initialize the NN with pre-trained weights from ImageNet~\cite{DBLP:journals/ijcv/RussakovskyDSKS15} 
As the $\vec{h}_{v(KG)}$ space of KG-NN depends on the KG instead of the source dataset, KG-NN can be applied to other domains following the same semantic relations given by the KG.
This property is shown on the domain generalization task.

\paragraph{Target Domain Pre-Training.}
Small amounts of labeled target data can usually be gathered with manageable effort.
However, just fine-tuning an NN with additional target domain data using the cross-entropy loss leads to catastrophic forgetting and thus poor accuracy.
We assume that this happens because the NN tries to find a new $\vec{h}_{v}$ that fits the target domain, but differs from the embedding obtained from the source domain.
In contrast, NNs optimized on the source domain using a KG as a trainer, can simply be enriched with additional target data using the same training method.
Therefore, KG-NN pre-trained on the source domain, is retrained on the target dataset using the same $\vec{h}_{KG}$. 

\paragraph{Linear Layer Training.}
For adaption to a downstream task like classification, we add a randomly-initialized linear fully-connected layer to the trained encoder network. 
The size of the output vector depends on the number of classes.
This linear layer is trained with the default cross-entropy loss, while the parameters of the encoder network remain unchanged.

\section{Experiment}
\label{sec:evaluation}
We conduct experiments on two different scenarios with multiple datasets to demonstrate the benefit of training an NN using a knowledge graph as a trainer, which leads to more accurate and more robust models in terms of the distribution shift.
We compare KG-NN with two baselines: 1) CE, which trains the NN using the supervised cross-entropy loss; and 2) SupCon, which trains the NN with the (self-)supervised contrastive loss~\cite{DBLP:conf/nips/KhoslaTWSTIMLK20}.
We chose CE, as it is typically used for training NNs, as well as SupCon, as this approach utilizes a similar contrastive loss function, however without the incorporation of prior knowledge and supervision.
CE and SupCon learn an embedding layer based on the data distribution of the source dataset, whereas KG-NN relies on the embedding given by the knowledge graph.
To qualitatively evaluate the influence of the knowledge graph embedding we further compare against GloVe, a variation of KG-NN that uses a GloVe~\cite{DBLP:conf/emnlp/PenningtonSM14} language embedding instead of $\vec{h}_{KG}$.
All approaches use the same ResNet-50~\cite{DBLP:conf/cvpr/HeZRS16} backend as encoder network and only differ in their method how this encoder network is trained.

Two different scenarios are defined to analyze our approach to concrete transfer learning tasks.
\emph{Scenario 1} - we investigate the sensitivity to distribution shifts using a domain generalization task. 
Therefore, we train: 
a) KG-NN, CE, SupCon, and GloVe from scratch on mini-ImageNet and evaluate on its derivatives, ImageNetV2~\cite{DBLP:conf/icml/RechtRSS19}, ImageNet-R~\cite{hendrycks2020many}, ImageNet-Sketch~\cite{wang2019learning} and ImageNet-A~\cite{hendrycks2019nae}; 
b) KG-NN, CE, and SupCon from scratch on GTSRB, and evaluate on CTSD.
\emph{Scenario 2} - we focus on supervised domain adaptation, a more practical scenario where KG-NN, CE, and SupCon are trained on GTSRB and fully retrained on CTSD with a small amount of target data.
Note that we exclude GloVe when using GTSRB/CTSD since the language embedding does not contain a specific representation for each roadsign class and therefore can not be applied straight forward.


\subsection{Scenario 1 - Domain Generalization}

Domain generalization describes the task of learning generalized models on a source domain so that they can be used on unseen target domains. 
Therefore, KG-NN is used without the target domain pre-training phase.

\subsubsection{Experiment 1 - Wordnet-Subset with mini-ImageNet}
\paragraph{Dataset Settings.}
As source domain, we use mini-ImageNet, a derivative of the ImageNet dataset, consisting of 60K color images of size 84 × 84 with 100 classes, each having 600 examples. 
Compared to ImageNet, this dataset fits in memory on modern machines, making it very convenient for rapid prototyping and experimentation.
For the evaluation, we use the target domains:
ImageNetV2, which contains 10 new test images per class and closely follows the original labeling protocol; 
ImageNet-R, which has art, cartoons, deviantart, etc. renditions of 200 ImageNet classes resulting in 30,000 images;
ImageNet-Sketch comprising 50,000 images, 50 images for each of the 1000 ImageNet classes; and \mbox{ImageNet-A}, which contains real-world, unmodified, and naturally occurring examples that cause machine learning model's performance to significantly degrade.

\paragraph{Knowledge Graph and KG Embedding Space.}
WordNet is a lexical database containing nouns, verbs, adjectives, and adverbs of the English language structured into respective synsets~\cite{DBLP:journals/cacm/Miller95,wordnet}. 
Each synset is an underlying concept consisting of a collection of synonyms as well as its relations to other synsets.
The \emph{Mini WordNet Knowledge Graph} (MWKG) is created by extracting the respective synsets of each label from the mini-ImageNet dataset from~\cite{wordnetRDF} into RDF representation.
These synsets are grouped based on the lexical domain they pertain to, e.g. \emph{animal}, \emph{artifact}, or \emph{food}.
They are represented as classes and further described with relations such as: \emph{hypernym}, \emph{meronym}, \emph{synset-member}.
Additionally, a shallow taxonomy is established by extracting the parents of each synset including their relationships and attributes.
In total, MWKG contains 198 classes with 8 annotation properties.
We transfer MWKG into a 300-dimensional $\vec{h}_{KG}$ using the MRGCN~\cite{DBLP:journals/corr/abs-2003-12383}, which exploits the literal information in addition to classes and their relationships.
To realize that, we use the MRGCN's node classification feature to build the $\vec{h}_{KG}$ that explicitly clusters the six lexical domains: animal, artifact, communication, food, object, and plant.

\paragraph{Training Details.}
All models use a ResNet-50 backend and are pre-trained with a batch size of 1024 on the mini-ImageNet dataset.
We resize the images to 32x32 for fast prototyping.
KG-NN and SupCon are trained for 1000 epochs using their respective contrastive loss function, stochastic gradient descent (SGD) with a learning rate of 0.5, cosine annealing, and a temperature of $\tau = 0.5$.
CE is trained for 500 epochs with the cross-entropy loss and SGD with a learning rate of 0.8.
For the \emph{linear-layer phase}, we train an \emph{one-layer} MLP on top of the frozen encoder networks of KG-NN, SupCon, and CE, with an adam optimizer and a learning rate of 0.0004.

\begin{figure}[tb]
    \centering
    \includegraphics[width=\textwidth]{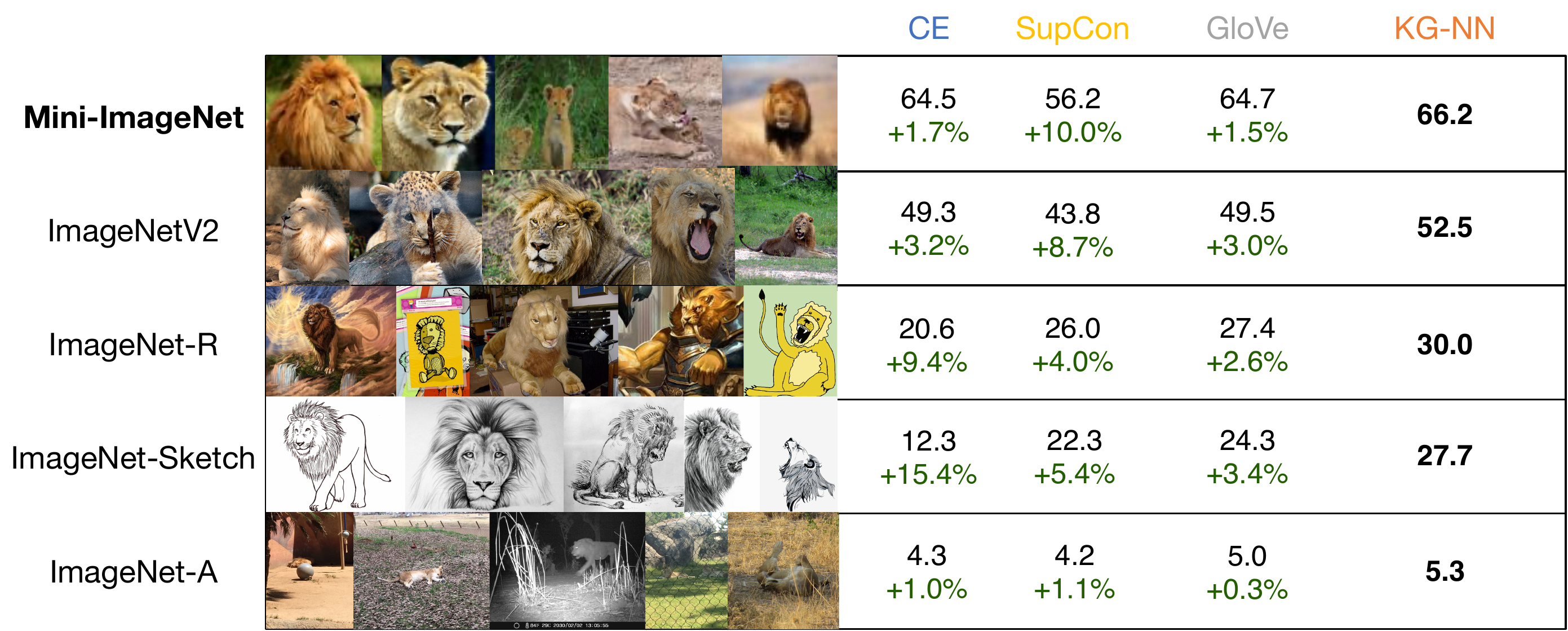}
    \caption{Accuracy of the domain generalization task using mini-ImageNet as source and multiple derivatives as target domains. We compare KG-NN with the standard CE, SupCon, a version of our loss without auxiliary knowledge of a KG, and GloVe, a version of KG-NN using a language embedding instead of a $\vec{h}_{KG}$.}
    \label{fig:mini-ImageNet experiment}

\end{figure}

\paragraph{Evaluation.}
We evaluate the models on ImageNetV2, ImageNet-R, ImageNet-Sketch, and ImageNet-A.
KG-NN outperforms CE, SupCon, and GloVe on the trained source as well as on unknown target domains as shown in Figure~\ref{fig:mini-ImageNet experiment}.
This means that KG-NN makes use of the additional semantic information.
It can be seen that CE fails particularly when the domain gap increases.
We assume that this happens due to its high specialization on the source domain.
SupCon cannot reach the performance of CE on the source dataset, however, it outperforms CE on more general target tasks.
We see that pre-training on a more generic self-supervised task helps the NN to extract more general features.
GloVe, the version of KG-NN that relies on a language embedding instead of a KG, is also outperformed by KG-NN.
We see that the performance of KG-NN depends on the quality of the embedding space, which we can control manually using different KGs or $KGE(\cdot)$s.

\subsubsection{Experiment 2 - RoadSign KG with GTSRB and CTSD}

\paragraph{Dataset Settings.}
The German Traffic Sign Dataset (GTSRB), which contains $51,970$ images of $43$ road signs, is used as the source domain, and the Chinese Traffic Sign Dataset (CTSD), which contains $6,164$ images of $58$ road signs, as the target domain.
We resize all images to a uniform size of $32x32$ pixels.
Note that we do not cut out the road signs, but take the whole image for classification.
Both datasets contain a domain shift as they were recorded with different cameras in different countries and hence have different appearances.

\paragraph{Knowledge Graph and KG Embedding Space.}
First, we construct a small knowledge graph for traffic sign recognition (RSKG) that contains all classes of both datasets incorporated in an underlying domain ontology.
To encode the formal semantics of road signs from different countries and standards, we first develop the \emph{RoadSign} ontology.
It contains classes (e.g. RoadSign, Shape, Icon, Color), relationships (e.g. hasShape, hasIcon, hasColor) and attributes (e.g. label, textWithinSign, etc).
The actual road signs that exist within given datasets are represented as concrete \emph{individuals}.
Note that this information is extracted from externally available road-sign standards, without accessing the datasets.
Currently, RSKG contains 18 classes, 11 object properties, 2 datatype properties, and 101 individuals. 
It is important to mention that the knowledge graph can be further populated with concrete road signs instances from other countries. 
This would enrich RSKG and could help to find inter-relations between the domains.
We transfer RSKG into a 300-dimensional $\vec{h}_{KG}$ by using MRGCN~\cite{DBLP:journals/corr/abs-2003-12383} as we also want to exploit its literal information.
Therefore, we use MRGCN in the node classification task to build a $\vec{h}_{KG}$ that explicitly clusters the five classes: danger, informative, mandatory, prohibitory, and warning.

\paragraph{Training Details.}
We use the same training setting and hyperparameters as in the experiment with the mini-ImageNet dataset.

\begin{figure}[tb]
    \centering
    \includegraphics[width=0.95\textwidth]{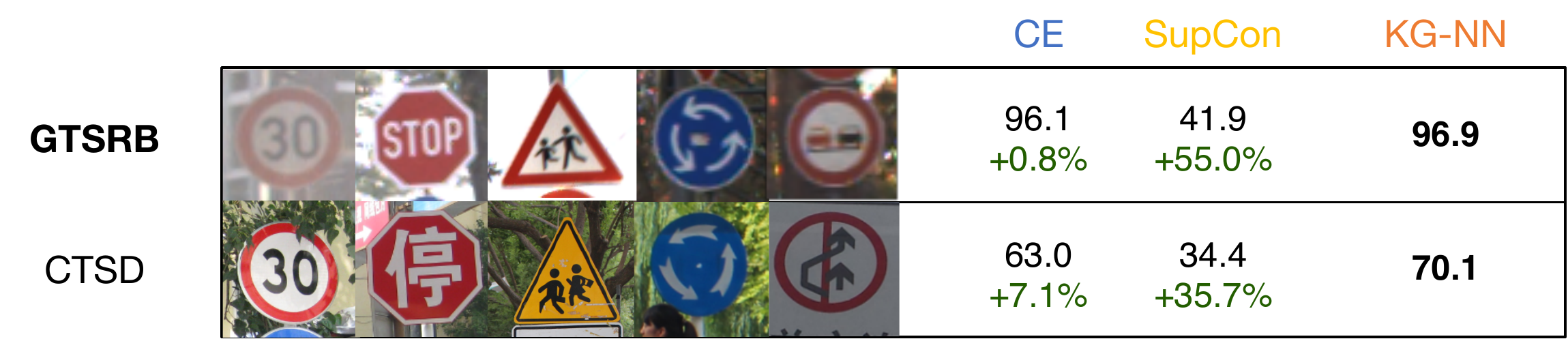}

    \caption{Accuracy of the domain generalization task using GTSRB as the source and CTSD as the target domain. We compare KG-NN with the standard CE and SupCon, a version of our loss without auxiliary knowledge of a KG.}
    \label{fig:gtsrb experiment}

\end{figure}

\paragraph{Evaluation.}
Figure~\ref{fig:gtsrb experiment} shows that KG-NN outperforms CE by 0.8\% on the source and by 7.1\% on the target dataset.
It can be seen that KG-NN exceeds the accuracy of SupCon by 55.0\% on GTSRB and by 35.7\% on CTSD.
SupCon with its self-supervised loss needs large datasets to form a good embedding space, however, both datasets are quite small and from the special domain of road-sign-recognition.
We do not compare against a GloVe embedding, as there are no instances for specific road signs and no clear procedure on how to generate these instances from a text corpus.
Overall, KG-NN performs better and is more robust to unforeseen distribution shifts using the same amount of training data.

\subsection{Scenario 2 - Supervised Domain Adaptation}
\begin{figure}[tb]
\centering
\begin{subfigure}{0.49\textwidth}
\centering
\includegraphics[width=\textwidth]{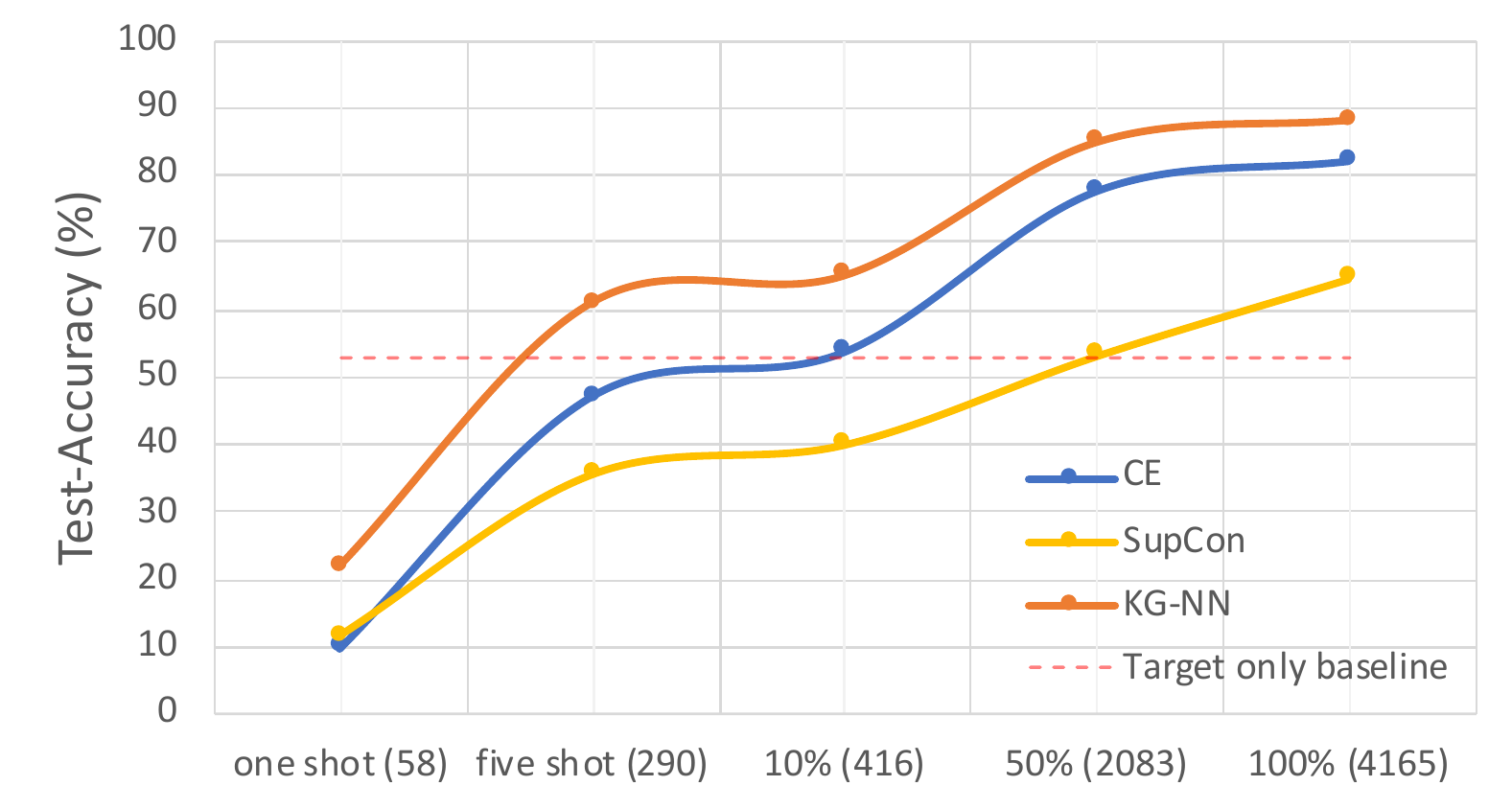}
\caption{Evaluation on CTSD}\label{fig:diagram_a}
\end{subfigure}
\begin{subfigure}{0.49\textwidth}
\includegraphics[width=\textwidth]{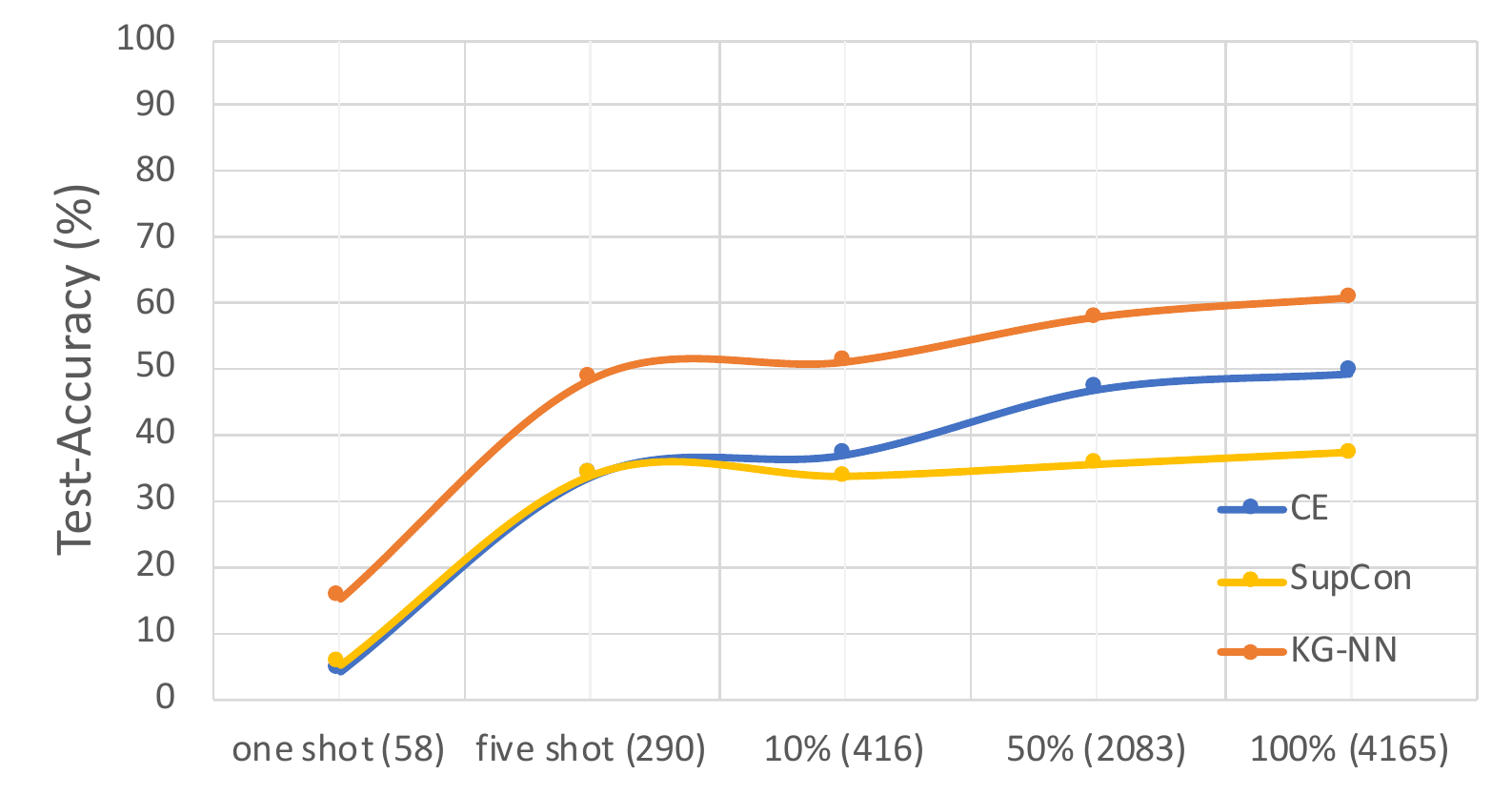}
\caption{Evaluation on GTSRB}\label{fig:diagram_b}
\end{subfigure}
\caption{Comparison of KG-NN, SupCon, and CE on the test dataset of the target domain for five different amounts of target data: a) evaluates the NNs on the target domain; b) evaluates the same NNs on the initial source domain to reflect \emph{catastrophic forgetting} phenomena.}
\label{fig:diagram}
\end{figure}

Supervised domain adaptation describes the task of transfer learning that adapts models learned on a source domain to a specific labeled target domain.
We claim that an NN learned using an image-data-independent $\vec{h}_{KG}$ can adapt to new domains and new classes better as both domains use the same embedding space.
For this experiment, we use the same settings described in Experiment 2. 
First, KG-NN, CE, and SupCon, are pre-trained on the source dataset.
Second, we use the encoder networks of each NN and presume the pre-training on the target dataset.
The NNs are retrained with different amounts of labeled target data.
The one-shot (58) experiment uses 58 images, one image for each class of the CTSD target dataset.
The five-shot (290) experiment uses 290 images, five images for each class of the CTSD.
The 10\% (416) experiment uses 416 images, 10\% of images of the CTSD.
The 50\% (2083) experiment uses 2,083 images, 50\% of images of the CTSD.
The 100\% (4165) experiment uses 4,165 images, 100\% of images of the CTSD target dataset.
Similar to the previous experiments, we use the \emph{linear layer phase} to adopt the pre-trained encoder network to the target task.
As shown in Figure~\ref{fig:diagram}, all experiments are evaluated on the full CTSD target dataset and on the 25 common classes of the GTSRB source dataset.

Evaluating the approaches on the initial source domain, we find that all NN suffer from \emph{catastrophic forgetting}, as depicted in Figure~\ref{fig:diagram_b}.
If 100\% of target data is used for training, the accuracy of CE drops from 96.1\% to 49.5\%, the accuracy of SupCon drops from 41.9\% to 37.2\%, and the accuracy of KG-NN drops from 96.9\% to 60.7\% on the source domain.
This means that KG-NN is still the best performing model on the source domain, even after retraining on a target domain with an increased difference to CE from 0.8\% to 11.2\%.
We think that the fixed embedding space between source and target domain helps to overcome the issue of \emph{catastrophic forgetting}.

If we compare the approaches on the target domain as illustrated in Figure~\ref{fig:diagram_a}, we see that KG-NN achieves an accuracy of 88.1\%, which is an improvement by 5.9\% over standard CE and by 23.4\% over SupCon.
Since we operate on transfer learning, an additional target-only baseline is introduced.
Thus, CE is initialized with weights pre-trained on ImageNet, instead of using the source domain to pre-train the parameters of the NN.
We see that the target-only baseline suffers from fewer target data in $D_T$ yielding only 53.1\% accuracy as the ImageNet initialization does not suit well for the task of road sign recognition.
All approaches seem to be able to transfer some knowledge from the source domain $D_S$ to the target domain $D_T$ outperforming the target-only baseline.
However, KG-NN significantly outperforms the baseline by 35.0\%, whereas CE improves by 29.1\% and SupCon by 11.6\%.

Interestingly, with less than five target images per class, which is fewer than 7\% of target data, KG-NN surpasses the performance of the target-only baseline.
We observe KG-NN always outperforms CE by approximately 10\% of accuracy.
When compared to SupCon, we see the accuracy difference even increases if more labeled target data is available.
In the one-shot scenario, KG-NN outperforms CE by 12.2\% of accuracy, in the five-shot-scenario by 13.8\%, in the 10\%-scenario by 11.2\%, in the 50\%-scenario by 10.7\%, and on the full target dataset by 5.9\%.
In the one-shot scenario KG-NN outperforms SupCon by 10.3\% of accuracy, in the five-shot-scenario by 25.4\%, in the 10\%-scenario by 25\%, in the 50\%-scenario by 31.6\%, and on the full target dataset by 23.4\%.

\section{Related Work}
\label{sec:related-work}
Embedding spaces trained with the cross-entropy loss tend to be specialized embedding spaces for a particular domain.
To reduce the high dependency on the training domain, pre-training methods that generate rich embedding spaces seem to be a promising research direction for CV and NLP.
Most neuro-symbolic approaches only learn a transformation function, e.g., MLP, on top of a pre-trained $\vec{h}_{v}$.
We refer to these models as visual-semantic transformation models.
Since the weights of the visual feature extractor are a really important part of robust object recognition, recent approaches have shown that learning a visual-semantic feature extractor from scratch improves generalization capabilities and makes the NN applicable to further downstream and transfer learning tasks~\cite{radford2learning}.
We refer to these models as visual-semantic feature extractors.


\paragraph{Neural Networks improved by Knowledge Graphs}
Most of the works that combine KGs with NNs use WordNet~\cite{Wang2018ZeroShotRV}, small-scale label~\cite{DBLP:conf/cvpr/LeeFYW18,DBLP:conf/cvpr/ChenWWG19} or scene~\cite{DBLP:conf/cvpr/ChenLFG18} graphs as KG.
However, the capacity of WordNet as a lexical database is limited.
Large-scale KGs such as DBPedia~\cite{DBLP:conf/semweb/AuerBKLCI07} or ConceptNet~\cite{DBLP:conf/aaai/SpeerCH17} encode additional semantic information by using higher order relations between concepts.
Although their applications are still sparse in the visual domain, there are a few works that have shown promising results.
DBPedia is already used in the field of explainable AI~\cite{DBLP:journals/corr/abs-1901-08547,DBLP:series/ssw/LecueCPC20}, object detection~\cite{DBLP:journals/corr/abs-1908-04385}, and visual question answering~\cite{DBLP:conf/ijcai/WangWSDH17}; and ConceptNet is used for video classification~\cite{DBLP:journals/corr/abs-1711-01714} and zero-shot action recognition~\cite{DBLP:conf/aaai/GaoZX19}.
However, all approaches use the KG only as a post-validation step on a pre-trained visual feature extractor, while KG-NN learns the visual feature extractor by itself based on the KG.

\paragraph{Visual-Semantic Transformation Models} are learned via a transformation function, e.g. MLP, from a pre-trained $\vec{h}_{v}$ into $\vec{h}_{s}$.
One of the first approaches that use $\vec{h}_{s}$ with NNs is the work from Mitchell et al.~\cite{Mitchell1191}. 
They use word embeddings derived from text corpus statistics to generate neural activity patterns, i.e. images.
Instead of generating images from text, Palatucci et al.~\cite{DBLP:conf/nips/PalatucciPHM09} learn a linear regression model to map neural activity patterns into word embedding space.
In their work, they improve zero-shot learning by extrapolating the knowledge gathered from in the $\vec{h}_{s}$ related classes to novel classes.
Socher et al.~\cite{DBLP:conf/nips/SocherGMN13} present a model for zero-shot learning that learns a transformation function between an $\vec{h}_{v}$ space, obtained by an unsupervised feature extraction method, and an $\vec{h}_{s}$, based on an NN-based language model.
The authors trained a 2-layer NN with the MSE loss to transform the $\vec{h}_{v}$ into the word embedding of 8 classes.
Frome et al.~\cite{DBLP:conf/nips/FromeCSBDRM13} introduce the deep visual-semantic embedding model DeViSE that extends the approach from 8 known and 2 unknown classes to 1000 known classes for the image model and up to 20,000 unknown classes.
Therefore, they pre-train their visual feature extractor using ImageNet and their $\vec{h}_{s}$ based on the Word2Vec~\cite{DBLP:conf/nips/MikolovSCCD13} language model, exposed to the text of a single online encyclopedia.
In contrast to Socher et al.~\cite{DBLP:conf/nips/SocherGMN13}, DeVISE learns a linear transformation function between the $\vec{h}_{v}$ space and the $\vec{h}_{s}$ space using a combination of dot-product similarity and hinge rank loss since MSE distance fails in high dimensional space.
Norouzi et al.~\cite{DBLP:journals/corr/NorouziMBSSFCD13} propose \emph{convex combination of semantic embeddings} (ConSE), a simple framework for constructing a zero-shot learning classifier.
ConSE uses a semantic word embedding model to reason about the predicted output scores of the NN-based image classifier.
To predict unknown classes, it performs a convex combination of the classes in the $\vec{h}_{s}$ space, weighted by their predicted output scores of the NN.
Similarly, Zhang et al.~\cite{DBLP:conf/iccv/ZhangS15a} introduce the \emph{semantic similarity embedding} (SSE), which models target data instances as a mixture of seen class proportions.
SSE builds a semantic space where each novel class could be represented as a probabilistic mixture of the projected source attribute vectors of the seen classes.
Akata et al.~\cite{DBLP:journals/pami/AkataPHS16} refer to their $\vec{h}_{s}$ space transformations as label embedding methods.
They compared transformation functions from the $\vec{h}_{v}$ space to the attribute label embedding space, the hierarchy label embedding space, and the Word2Vec label embedding space, in which embedded classes can share features among themselves.

\paragraph{Visual-Semantic Feature Extractors:}
The approaches mentioned so far only learn a transformation from $\vec{h}_{v}$ to $\vec{h}_{s}$. 
However, the parameters of the feature extractor are not affected by the auxiliary information.
Thus, if the feature extractor cannot detect visual features due to the domain shift problem, the performance of the final prediction suffers.
Instead of maximizing the likelihood on the output, some approaches maximize the energy (i.e. difference between the prediction and the excepted result) directly on the embedding space to learn the NN.
Hadsell et al.~\cite{Hadsell2006DimensionalityRB} introduce the contrastive loss for a \emph{siamese architecture} to learn a robust embedding space from unlabeled data.
They show that their self-supervised energy-based method can learn a lighting and rotation-invariant embedding space.
Recently, many approaches claim that training an embedding space in a self-supervised manner using the contrastive loss tends to find a more general and domain-invariant representation~\cite{DBLP:conf/icml/ChenK0H20,He2020MomentumCF}.
Furthermore, Tian et al.~\cite{Tian2020RethinkingFI} show that learning an embedding space using the contrastive loss, followed by training a supervised linear classifier on top of this representation, outperforms state-of-the-art few-shot learning methods.

Joulin et al.~\cite{DBLP:conf/eccv/JoulinMJV16} demonstrate that feature extractors trained to predict words in image captions can learn useful visual-semantic embedding spaces $\vec{h}_{v(s)}$.
Further, Radford et al.~\cite{radford2learning} proposed a simple and general pre-training of an NN with natural language supervision using a dataset of 400 million image-text pairs collected from the internet and the contrastive objective of Zhang et al.~\cite{DBLP:journals/corr/abs-2010-00747}.

To the best of our knowledge, there is no work that learn a visual feature extractor using a KG or its embedding space $\vec{h}_{KG}$.
We choose to use prior knowledge encoded in a knowledge graph instead of using the unstructured knowledge of a language embedding as they are highly dependent on their text corpus, inconsistent, and do not incorporate expert knowledge.

\section{Conclusion and Future Work}
\label{sec:conclusion}



In this paper, we propose KG-NN, a knowledge graph-based approach that enables NN to learn more robust and controlled embedding spaces for transfer learning tasks.
The core idea of our approach is to use domain-invariant knowledge represented in a KG, transform it into a vector space using knowledge graph embedding algorithms, and train an NN so that its embedding space is adapted to the domain-invariant embeddings given by the KG.
Using our KG-based contrastive loss function, we force the NN to adapt its $\vec{h}_{v}$ space to the domain-invariant space $\vec{h}_{KG}$ given by the KG, thus forming $\vec{h}_{v(KG)}$.
Our experimental results show that NNs benefit from exploiting prior knowledge.
As a result, it increases the accuracy on known and unknown domains and allows them to keep up with NNs trained with the cross-entropy loss despite requiring significantly less training data.

There are several directions of future work: 
First, identifying discriminative factors to best influence the domain-invariant space.
Therefore, further investigations are needed to determine \email{what} knowledge is relevant and should be modeled in the KG to enable transfer learning.
Second, analyzing \email{how} the prior knowledge can be modeled and represented best, e.g., via n-ary relations or hyper-relational graphs.
Third, exploring various embedding techniques to operate on multi-modal information or Riemannian metrics to exploit hierarchical relations.
And finally, evaluating different contrasting dimensions and knowledge infusion techniques 
could lead to further improvements.

We believe that the construction of task-specific knowledge graph embeddings and their combination with learned embeddings of NNs will help to build more interpretable, more robust, and more accurate machine learning models, while at the same time requiring less training data.

\section{Acknowledgement}

This publication was created as part of the research project "KI Delta Learning" (project number: 19A19013D) funded by the Federal Ministry for Economic Affairs and Energy (BMWi) on the basis of a decision by the German Bundestag.


\bibliographystyle{splncs04}
\bibliography{literature}

\begin{thebibliography}{10}
\providecommand{\url}[1]{\texttt{#1}}
\providecommand{\urlprefix}{URL }
\providecommand{\doi}[1]{https://doi.org/#1}

\bibitem{DBLP:journals/pami/AkataPHS16}
Akata, Z., Perronnin, F., Harchaoui, Z., Schmid, C.: Label-embedding for image
  classification. {IEEE} Trans. Pattern Anal. Mach. Intell.  (2016)

\bibitem{wordnetRDF}
van Assem, M., Isaac, A., von Ossenbruggen, J.: Word{N}et 3.0 in {RDF} (2010),
  \url{http://semanticweb.cs.vu.nl/lod/wn30/}

\bibitem{DBLP:conf/semweb/AuerBKLCI07}
Auer, S., Bizer, C., Kobilarov, G., Lehmann, J., Cyganiak, R., Ives, Z.G.:
  Dbpedia: {A} nucleus for a web of open data. In: ISWC (2007)

\bibitem{DBLP:conf/icml/ChenK0H20}
Chen, T., Kornblith, S., Norouzi, M., Hinton, G.E.: A simple framework for
  contrastive learning of visual representations. In: ICML (2020)

\bibitem{DBLP:conf/cvpr/ChenLFG18}
Chen, X., Li, L., Fei{-}Fei, L., Gupta, A.: Iterative visual reasoning beyond
  convolutions. In: CVPR (2018)

\bibitem{DBLP:conf/cvpr/ChenWWG19}
Chen, Z., Wei, X., Wang, P., Guo, Y.: Multi-label image recognition with graph
  convolutional networks. In: CVPR (2019)

\bibitem{DAmour2020UnderspecificationPC}
D'Amour, A., Heller, K.A., Moldovan, D., Adlam, B., Alipanahi, B., Beutel, A.,
  Chen, C., Deaton, J.: Underspecification presents challenges for credibility
  in modern machine learning. CoRR  (2020)

\bibitem{DBLP:conf/aaai/DettmersMS018}
Dettmers, T., Minervini, P., Stenetorp, P., Riedel, S.: Convolutional 2d
  knowledge graph embeddings. In: AAAI (2018)

\bibitem{DBLP:conf/nips/FromeCSBDRM13}
Frome, A., Corrado, G.S., Shlens, J., Bengio, S., Dean, J., Ranzato, M.,
  Mikolov, T.: Devise: {A} deep visual-semantic embedding model. In: NIPS
  (2013)

\bibitem{DBLP:conf/aaai/GaoZX19}
Gao, J., Zhang, T., Xu, C.: I know the relationships: Zero-shot action
  recognition via two-stream graph convolutional networks and knowledge graphs.
  In: AAAI (2019)

\bibitem{DBLP:journals/corr/abs-1901-08547}
Geng, Y., Chen, J., Jim{\'{e}}nez{-}Ruiz, E., Chen, H.: Human-centric transfer
  learning explanation via knowledge graph [extended abstract]. CoRR  (2019)

\bibitem{DBLP:books/daglib/0040158}
Goodfellow, I.J., Bengio, Y., Courville, A.C.: Deep Learning. Adaptive
  computation and machine learning (2016)

\bibitem{Goodfellow2015ExplainingAH}
Goodfellow, I.J., Shlens, J., Szegedy, C.: Explaining and harnessing
  adversarial examples. In: Bengio, Y., LeCun, Y. (eds.) ICLR (2015)

\bibitem{Hadsell2006DimensionalityRB}
Hadsell, R., Chopra, S., LeCun, Y.: Dimensionality reduction by learning an
  invariant mapping. In: CVPR (2006)

\bibitem{DBLP:books/sp/HastieFT01}
Hastie, T., Friedman, J.H., Tibshirani, R.: The Elements of Statistical
  Learning: Data Mining, Inference, and Prediction (2001)

\bibitem{He2020MomentumCF}
He, K., Fan, H., Wu, Y., Xie, S., Girshick, R.B.: Momentum contrast for
  unsupervised visual representation learning. In: CVPR (2020)

\bibitem{DBLP:conf/cvpr/HeZRS16}
He, K., Zhang, X., Ren, S., Sun, J.: Deep residual learning for image
  recognition. In: CVPR (2016)

\bibitem{hendrycks2020many}
Hendrycks, D., Basart, S., Mu, N., Kadavath, S., Wang, F., et~al., E.D.: The
  many faces of robustness: A critical analysis of out-of-distribution
  generalization. CoRR  (2020)

\bibitem{Hendrycks2019BenchmarkingNN}
Hendrycks, D., Dietterich, T.G.: Benchmarking neural network robustness to
  common corruptions and perturbations. In: ICLR (2019)

\bibitem{hendrycks2019nae}
Hendrycks, D., Zhao, K., Basart, S., Steinhardt, J., Song, D.: Natural
  adversarial examples. CoRR  (2019)

\bibitem{DBLP:journals/corr/abs-2003-02320}
Hogan, A., Blomqvist, E., Cochez, M., d'Amato, C., de~Melo, G., Gutierrez, C.,
  Gayo, J.E.L., et~al., S.K.: Knowledge graphs. CoRR  (2020)

\bibitem{DBLP:conf/eccv/JoulinMJV16}
Joulin, A., van~der Maaten, L., Jabri, A., Vasilache, N.: Learning visual
  features from large weakly supervised data. In: {ECCV}. Springer (2016)

\bibitem{DBLP:conf/nips/KhoslaTWSTIMLK20}
Khosla, P., Teterwak, P., Wang, C., Sarna, A., Tian, Y., Isola, P., Maschinot,
  A., Liu, C., Krishnan, D.: Supervised contrastive learning. In: NeurIPS
  (2020)

\bibitem{DBLP:series/ssw/LecueCPC20}
L{\'{e}}cu{\'{e}}, F., Chen, J., Pan, J.Z., Chen, H.: Knowledge-based
  explanations for transfer learning. Studies on the Semantic Web (2020)

\bibitem{DBLP:conf/cvpr/LeeFYW18}
Lee, C., Fang, W., Yeh, C., Wang, Y.F.: Multi-label zero-shot learning with
  structured knowledge graphs. In: CVPR (2018)

\bibitem{DBLP:journals/corr/abs-1908-04385}
Liu, Z., Jiang, Z., Wei, F.: {OD-GCN} object detection by knowledge graph with
  {GCN}. CoRR  (2019)

\bibitem{DBLP:conf/nips/MikolovSCCD13}
Mikolov, T., Sutskever, I., Chen, K., Corrado, G.S., Dean, J.: Distributed
  representations of words and phrases and their compositionality. In: NIPS
  (2013)

\bibitem{DBLP:journals/cacm/Miller95}
Miller, G.A.: Wordnet: {A} lexical database for english. Commun. {ACM}  (1995)

\bibitem{Mitchell1191}
Mitchell, T.M., Shinkareva, S.V., Carlson, A., Chang, K.M., Malave, V.L.,
  Mason, R.A., Just, M.A.: Predicting human brain activity associated with the
  meanings of nouns. Science  (2008)

\bibitem{DBLP:conf/naacl/NguyenNNP18}
Nguyen, D.Q., Nguyen, T.D., Nguyen, D.Q., Phung, D.Q.: A novel embedding model
  for knowledge base completion based on convolutional neural network. In:
  NAACL-HLT (2018)

\bibitem{Nickel2016HolographicEO}
Nickel, M., Rosasco, L., Poggio, T.A.: Holographic embeddings of knowledge
  graphs. In: Schuurmans, D., Wellman, M.P. (eds.) AAAI (2016)

\bibitem{DBLP:journals/corr/NorouziMBSSFCD13}
Norouzi, M., Mikolov, T., Bengio, S., Singer, Y., Shlens, J., Frome, A.,
  Corrado, G., Dean, J.: Zero-shot learning by convex combination of semantic
  embeddings. In: {ICLR} (2014)

\bibitem{DBLP:journals/corr/abs-1807-03748}
van~den Oord, A., Li, Y., Vinyals, O.: Representation learning with contrastive
  predictive coding. CoRR  (2018)

\bibitem{DBLP:conf/nips/PalatucciPHM09}
Palatucci, M., Pomerleau, D., Hinton, G.E., Mitchell, T.M.: Zero-shot learning
  with semantic output codes. In: NIPS (2009)

\bibitem{DBLP:conf/emnlp/PenningtonSM14}
Pennington, J., Socher, R., Manning, C.D.: Glove: Global vectors for word
  representation. In: EMNLP (2014)

\bibitem{radford2learning}
Radford, A., Kim, J.W., Hallacy, C., Ramesh, A., Goh, G., Agarwal, S., Sastry,
  G., Askell, A., Mishkin, P., Clark, J., et~al.: Learning transferable visual
  models from natural language supervision. Image  (2021)

\bibitem{DBLP:conf/icml/RechtRSS19}
Recht, B., Roelofs, R., Schmidt, L., Shankar, V.: Do imagenet classifiers
  generalize to imagenet? In: ICML (2019)

\bibitem{DBLP:conf/emnlp/RuderP17}
Ruder, S., Plank, B.: Learning to select data for transfer learning with
  bayesian optimization. In: EMNLP (2017)

\bibitem{DBLP:journals/ijcv/RussakovskyDSKS15}
Russakovsky, O., Deng, J., Su, H., et~al., J.K.: Imagenet large scale visual
  recognition challenge. Int. J. Comput. Vis.  (2015)

\bibitem{DBLP:conf/nips/SocherGMN13}
Socher, R., Ganjoo, M., Manning, C.D., Ng, A.Y.: Zero-shot learning through
  cross-modal transfer. In: NIPS (2013)

\bibitem{DBLP:conf/nips/Sohn16}
Sohn, K.: Improved deep metric learning with multi-class n-pair loss objective.
  In: NIPS (2016)

\bibitem{DBLP:conf/aaai/SpeerCH17}
Speer, R., Chin, J., Havasi, C.: Conceptnet 5.5: An open multilingual graph of
  general knowledge. In: AAAI (2017)

\bibitem{Stallkamp2012ManVC}
Stallkamp, J., Schlipsing, M., Salmen, J., Igel, C.: Man vs. computer:
  Benchmarking machine learning algorithms for traffic sign recognition. Neural
  Networks  (2012)

\bibitem{tan2018survey}
Tan, C., Sun, F., Kong, T., Zhang, W., Yang, C., Liu, C.: A survey on deep
  transfer learning. In: ICANN (2018)

\bibitem{Tian2020RethinkingFI}
Tian, Y., Wang, Y., Krishnan, D., Tenenbaum, J.B., Isola, P.: Rethinking
  few-shot image classification: {A} good embedding is all you need? In: ECCV
  (2020)

\bibitem{wordnet}
University, P.: About WordNet (2010), \url{https://wordnet.princeton.edu}

\bibitem{DBLP:conf/nips/VinyalsBLKW16}
Vinyals, O., Blundell, C., Lillicrap, T., Kavukcuoglu, K., Wierstra, D.:
  Matching networks for one shot learning. In: NIPS (2016)

\bibitem{wang2019learning}
Wang, H., Ge, S., Lipton, Z., Xing, E.P.: Learning robust global
  representations by penalizing local predictive power. In: NeurIPS (2019)

\bibitem{DBLP:conf/ijcai/WangWSDH17}
Wang, P., Wu, Q., Shen, C., Dick, A.R., van~den Hengel, A.: Explicit
  knowledge-based reasoning for visual question answering. In: IJCAI (2017)

\bibitem{Wang2018ZeroShotRV}
Wang, X., Ye, Y., Gupta, A.: Zero-shot recognition via semantic embeddings and
  knowledge graphs. In: CVPR (2018)

\bibitem{DBLP:journals/corr/abs-2003-12383}
Wilcke, W.X., Bloem, P., de~Boer, V., van~t Veer, R.H., van Harmelen, F.A.H.:
  End-to-end entity classification on multimodal knowledge graphs. CoRR  (2020)

\bibitem{Yang2016TowardsRT}
Yang, Y., Luo, H., Xu, H., Wu, F.: Towards real-time traffic sign detection and
  classification. {IEEE} Trans. Intell. Transp. Syst.  (2016)

\bibitem{DBLP:journals/corr/abs-1711-01714}
Yuan, F., Wang, Z., Lin, J., D'Haro, L.F., Jae, K.J., Zeng, Z., Chandrasekhar,
  V.: End-to-end video classification with knowledge graphs. CoRR  (2017)

\bibitem{DBLP:journals/corr/abs-2010-00747}
Zhang, Y., Jiang, H., Miura, Y., Manning, C.D., Langlotz, C.P.: Contrastive
  learning of medical visual representations from paired images and text. CoRR
  (2020)

\bibitem{DBLP:conf/iccv/ZhangS15a}
Zhang, Z., Saligrama, V.: Zero-shot learning via semantic similarity embedding.
  In: ICCV (2015)

\end{thebibliography}

\end{document}